\documentclass[conference]{IEEEtran}
\IEEEoverridecommandlockouts
\usepackage{cite}
\usepackage{amsmath,amssymb,amsfonts}
\usepackage{algorithmic}
\usepackage{url}
\usepackage{graphicx}
\usepackage{textcomp}
\usepackage{xcolor}
\usepackage{caption}
\usepackage{subcaption}
\def\BibTeX{{\rm B\kern-.05em{\sc i\kern-.025em b}\kern-.08em
    T\kern-.1667em\lower.7ex\hbox{E}\kern-.125emX}}
\begin{document}

\makeatletter
\newcommand{\linebreakand}{%
  \end{@IEEEauthorhalign}
  \hfill\mbox{}\par
  \mbox{}\hfill\begin{@IEEEauthorhalign}
}
\makeatother
    
\title{Farmer's Assistant: A Machine Learning Based Application for Agricultural Solutions\\
}

\author{\IEEEauthorblockN{Shloka Gupta}
\IEEEauthorblockA{\textit{Dept. of Information Technology} \\
\textit{Datta Meghe College of Engineering}\\
Navi Mumbai, India \\
shlokaprincess101@gmail.com}
\and
\IEEEauthorblockN{Nishit Jain}
\IEEEauthorblockA{\textit{Dept. of Information Technology} \\
\textit{Datta Meghe College of Engineering}\\
Navi Mumbai, India \\
yonishitjain123456@gmail.com}
\and
\IEEEauthorblockN{Akshay Chopade}
\IEEEauthorblockA{\textit{Dept. of Information Technology} \\
\textit{Datta Meghe College of Engineering}\\
Navi Mumbai, India \\
cakshay001@gmail.com}
\linebreakand 
\IEEEauthorblockN{Aparna Bhonde}
\IEEEauthorblockA{\textit{Dept. of Information Technology} \\
\textit{Datta Meghe College of Engineering}\\
Navi Mumbai, India \\
aparna.bhonde@dmce.ac.in}
}

\maketitle

\begin{abstract}
Farmers face several challenges when growing crops like uncertain irrigation, poor soil quality, etc. Especially in India, a major fraction of farmers do not have the knowledge to select appropriate crops and fertilizers. Moreover, crop failure due to disease causes a significant loss to the farmers, as well as the consumers. While there have been recent developments in the automated detection of these diseases using Machine Learning techniques, the utilization of Deep Learning has not been fully explored. Additionally, such models are not easy to use because of the high-quality data used in their training, lack of computational power, and poor generalizability of the models. To this end, we create an open-source easy-to-use web application to address some of these issues which may help improve crop production. In particular, we support crop recommendation, fertilizer recommendation, plant disease prediction, and an interactive news-feed. In addition, we also use interpretability techniques in an attempt to explain the prediction made by our disease detection model.
\end{abstract}

\begin{IEEEkeywords}
Leaf Image Analysis, Plant Disease Detection, Image Analysis, Deep Learning, Machine Learning, Interpretability
\end{IEEEkeywords}

\section{Introduction}
Agriculture is an extremely risky industry and our farmers are at the forefront of the industry. Farmers face several problems which include crops get affected by diseases, the soil is not being nutritious enough for the plant to grow, etc. All these factors reduce the overall yield. In India, more than 70\% population is dependent on agriculture \cite{pdduip}. More than 15\% of the crops get wasted in India due to diseases and hence it has become one of the major concerns to be resolved \cite{imagebased}.

Crop diseases are a major threat to crops, but their rapid identification remains difficult in many parts of the world due to the lack of the necessary infrastructure. The combination of increasing global smartphone penetration in the rural areas and recent advances in computer vision made possible by deep learning has paved the way for web-assisted disease diagnosis. There are several ways to detect disease for a plant such as growth, root, and leaves. A lot of existing systems use mobile applications to detect disease with the help of leaf. However, these applications work on leaf images with a flat and only black background. In addition, selecting proper fertilizers and crops according to the soil is of great significance, as it helps the farmers increase their crop yield, and improve their sales.

The three macro-nutrients used by plants are nitrogen, Phosphorus, and Potassium. N is responsible for the growth of leaves in a plant, P for root, flower and fruit, and K helps the overall function of the plant. Knowing the NPK values of fertilizer can help a farmer select a fertilizer that is appropriate for the type of plant they are growing.

In this paper, we propose a system which helps farmers detect plant disease, recommend the ideal crop for their soil and recommend fertilizers for them to get the best yield possible. We use the EfficientNet\cite{resnet} deep learning model, which achieves 99.8\% validation accuracy on our choice of dataset for plant disease detection, Random Forest model for crop recommendation based on the soil (N, P, K, pH) and weather features, and a rule-based classification system for fertilizer recommendation. In addition, we also provide them with news related to plants by scraping news websites, and perform explanations of our disease detection model with a popular interpretability technique called LIME\cite{lime} to understand our model better.

\section{Literature Survey}
\subsection{Deep Learning in Computer Vision}
Deep learning in computer vision has seen significant advancements, especially with the creation of the ImageNet\cite{imagenet} dataset and the ILSVRC\cite{ilsvrc} challenge. ImageNet is a popular dataset for pre-training deep learning models, which is currently the conventional way of handling computer vision problems with lack of data. ImageNet aims to populate the majority of the 80,000 synsets of WordNet\cite{wordnet} with an average of 500–1000 clean and full resolution images. Since its beginning, several deep convolutional neural networks have been designed to tackle the challenge. AlexNet\cite{alexnet} has 5 convolutional layers, whereas the VGG\cite{vgg} network has 19 layers. The introduction of ResNet\cite{resnet} took care of the Vanishing/Exploding gradient problem using residual connections. MobileNet\cite{mobilenet} is mindful of the restricted resources and is designed for mobile and embedded vision applications. EfficientNet\cite{efficientnet} proposes a model scaling method that uses a highly effective compound coefficient to scale up CNNs in a more structured manner.

\subsection{Plant Disease Detection}
Plant Disease Detection has been a very active field of research and there are several different techniques which have been proposed over the years, the latest ones using deep learning approaches.

In \cite{pdduip} for example, traditional image processing techniques such as noise removal, region cropping, image segmentation using boundary detection and Otsu thresholding were used. Features like color, texture and morphology were extracted and passed to an Artifical Neural Network.

The paper \cite{imagebased} used AlexNet and GoogLeNet with and without transfer learning on the PlantVillage dataset to achieve 99.35\% accuracy.  They also visualize activations and test on scraped data from Bing and Google Search.

Reference \cite{pdddlsurvey}  provides an excellent review of over hundred papers which use Deep Learning for plant disease detection and classification. They point out that a majority of the papers use PlantVillage dataset for their task, and deploy ImageNet-based pre-trained models (VGG, ResNet, Inception, DenseNet, etc.) as their model backbones. They also mention several visualization techniques used for this task, like - heatmaps, saliency maps, feature maps, activation visualization, segmentation maps, etc.

In \cite{imagebased}, the authors use VGG, ResNet, Inception-V3 on an augmented version of PlantVillage dataset with 87K images, and conclude that VGG is the best for their settings.

Another review paper \cite{pdpdsurvey} explores several papers from 2014 to 2020 which perform classification and detection of plant diseases and pests. They also point out the deep learning theory, classification after selecting region of interest methods, etc.

Reference \cite{classicaltodeep} discusses various classical machine learning and deep learning techniques used in detection of plant diseases. They also elaborate on how while there are several mobile and online applications for this task, only few of them are publicly available and accessible online. Moreover, they point out that these applications generalize poorly on real-life images, which have several leaves in the image and highly heterogeneous backgrounds.

Reference \cite{comparative} studies the performance of four popular deep learning models: VGG-16, ResNet-50, InceptionV4, DenseNet-121 on the PlantVillage datasets.

\subsection{Crop Recommendation}
In \cite{cropreco1}, the authors use - Random forests, Artificial Neural Nets, Support Vector Machines, etc. and conclude that Random forests work best for their dataset in crop recommendation. They also create a mobile application system which takes in location data using GPS and predicts the crop yield for a given crop, in addition to recommending crops based on area and soil quality as input. Similarly, \cite{cropreco2} uses a majority voting
on an ensemble of CHAID, Naive Bayes, K-NN and Random Trees for crop recommendation.

\subsection{Fertilizer Recommendation}
A lot of research has been done in fertilizer recommendation and a majority of them \cite{fertreco1, fertreco2, fertreco3, fertreco4} use the N, P, K, pH values of soil sometimes in addition with depth, temperature, weather, location, precipitation. The usual approach is to use rule based classification, but some approaches\cite{fertreco1} also use clustering on fertilizer data using K-Means and Random Forests for recommendation.

\subsection{Interpretability in Deep Learning}
The LIME\cite{lime} approach is a simple interpretability technique which uses a local linear regression surrogate for the original model. The linear model is trained on original model predictions on masked versions of the image. The scores for the image segments are based on their corresponding weights. Positive and higher scoring segments are important towards the predicted class, while the lower scoring ones hurt the confidence of the model.

The GradCAM\cite{gradcam} uses the average of gradients at the last convolutional layer of a CNN-based model to weight the activation maps at that layer, and performs a linear combination to find the postivey influence regions for a particular class. This provides a coarse heatmap of important regions towards a prediction.

\section{Methodology}
The following subsections describe our application and the machine learning involved in our experiments, with the implementation details as well as dataset and training information.
First, we describe our application with the help of a flowchart, as well as block diagrams which explain how the application user-interface is designed.
Then, we move on to our Machine Learning experiments, where we describe various models that we use, and other experimental details.
Both the subsections are further split into nested subsections, namely - crop recommendation, fertilizer recommendation, and plant diseases detection.
The machine learning subsection also includes an explanation of how we use LIME for interpretation, while the application subsection deals with the news feed implementation description.
\subsection{The Application}

\begin{figure}[htbp]
\centerline{\includegraphics[width=0.5\textwidth]{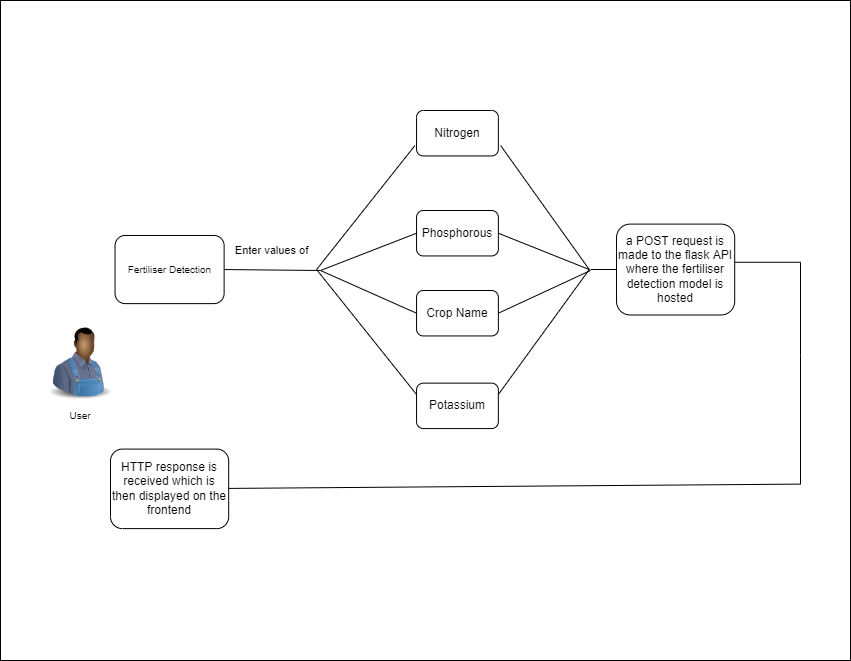}}
\caption{Flow diagram for Fertilizer Recommendation system}
\label{fig: flow_fertilizer_recommendation}
\end{figure}

\subsubsection{Fertilizer Recommendation}
The user has to enter the Nitrogen, Phosphorus, Potassium values along with the crop Name. A POST request is made to the flask API. Over here the fertilizer recommendation classifier is hosted. An HTTP response is sent to the front-end and in turn on the front-end the user gets recommendation to fertilizer. Fig.~\ref{fig: flow_fertilizer_recommendation} shows the flow diagram for the same.

\subsubsection{Disease Detection}
In disease detection the user has to click an image or directly upload it. The image is sent to the back-end and is processed by the model. After the image has been processed an HTTP response is sent to the front-end. The user receives the disease the plant has and its remedies. Fig.~\ref{fig: flow_plant_disease_detection} shows the flow diagram for the same.

\subsubsection{Crop Recommendation}
On entering the values of Nitrogen, phosphorus, and Potassium a post request is made to the flask API. After the model runs an HTTP response is sent to the front-end which tells the best crop a farmer can grow in the soil in order to get the best out of the land. Fig.~\ref{fig: flow_crop_recommendation} shows the flow diagram for the same. 

\subsubsection{News Feed}
The NDTV website has been scraped and then deployed as an API, when the user navigates to the news section of our website an HTTP response is sent to the front-end which displays latest news related to agriculture along with the link directed to the main article.
\begin{figure}[htbp]
\centerline{\includegraphics[width=0.5\textwidth]{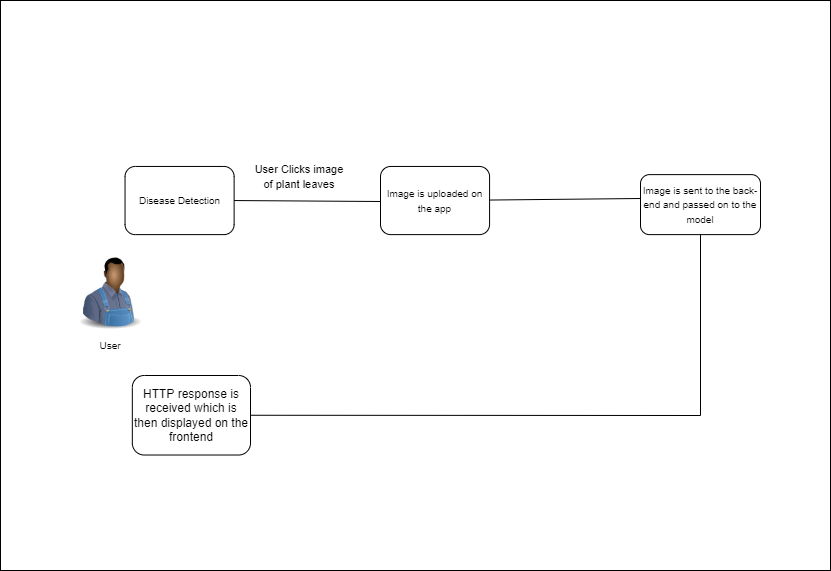}}
\caption{Flow diagram for Plant Disease Detection system}
\label{fig: flow_plant_disease_detection}
\end{figure}

\begin{figure}[htbp]
\centerline{\includegraphics[width=0.5\textwidth]{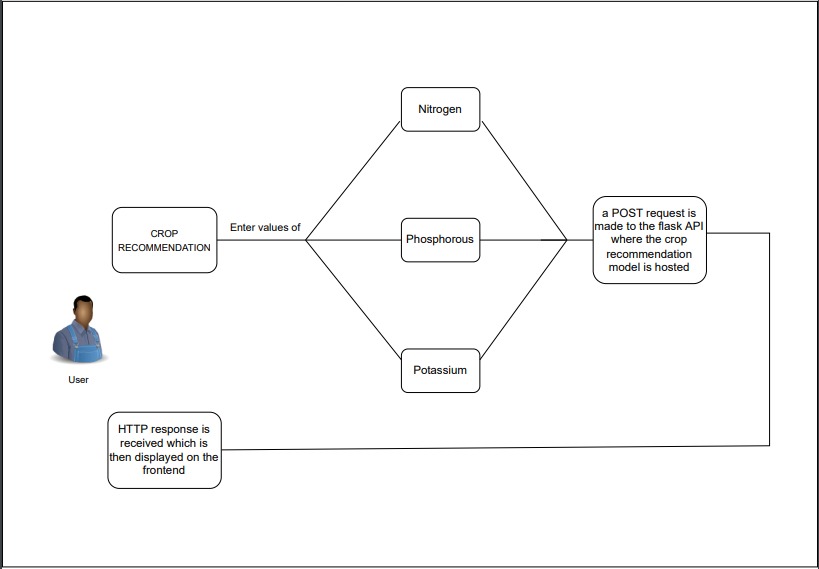}}
\caption{Flow diagram for Crop Recommendation system}
\label{fig: flow_crop_recommendation}
\end{figure}

\subsubsection{Disease Portal}
The disease portal provides a detailed view of various plant diseases and the kinds of products that may be bought to cure the plants of the disease.

\subsubsection{Interpretability Analysis}
The plant leaf image uploaded by the user is sent to a deployed API where the LIME computation takes place, the computing is done on a droplet server hosted on DigitalOcean which sends the resultant image in form of URI which is then displayed on the front-end.

\subsection{Machine Learning}

\subsubsection{Crop Recommendation}
\textit{Dataset Description:} This dataset is relatively simple with very few but useful features unlike the complicated features affecting the yield of the crop and has been taken from Kaggle
\footnote{Crop Recommendation Dataset: \url{https://www.kaggle.com/atharvaingle/crop-recommendation-dataset}}. It consists of 7 features namely - N: Ratio of Nitrogen content in the soil, P: Ratio of Phosphorus content in the soil, K:  ratio of Potassium content in the soil, temperature: Temperature in degree Celsius, Humidity: relative humidity in \%, ph: ph value of the soil, rainfall: rainfall in mm. The task is to predict the type of crop using these 7 features. The number of samples is 2200, and the total number of class labels are 22, some of which are:  rice, maize, coffee, muskmelon, etc. The number of samples per class is 100, which shows that the dataset is perfectly balanced and does not need any special imbalance handling technique.\\
\textit{Approach:} The dataset is split into 5-folds and cross-validation is performed on these folds. We test performance with six models:
\begin{itemize}
    \item Decision Tree with entropy as the criteria and a max depth of 5.
    \item Naive Bayes.
    \item SVM with a 0-1 scaling on the input, polynomial kernel with degree 3, and the L2 regularization parameter C=3.
    \item Logistic Regression.
    \item Random Forest with 20 estimators.
    \item XGBoost.
\end{itemize}

We use the sklearn library for all the models, except for XGBoost which is taken from the xgboost library. The parameters not mentioned are set to default for the purpose of our training. We select the best performing model among these for the application, and perform inference using the same.

\subsubsection{Disease Detection}
\textit{Dataset Description:}
For leaf disease detection, we consider the PlantVillage dataset. Specifically, we use an augmented version of the PlantVillage dataset present on Kaggle\footnote{Augmented PlantVillage Dataset: \url{https://www.kaggle.com/vipoooool/new-plant-diseases-dataset}}. The dataset contains 87,000 RGB examples of healthy and diseased crops, which have a spread of 38 class labels assigned to them. The number of crops included is 14, with a total of 26 different diseases. On average, each class contains ~1850 image samples with a standard deviation of ~104. The dataset has been split into 80:20 training to validation ratio. We attempt to predict the crop-disease pair given just the image of the plant leaf. We scale the images down by a factor of 255, and resize them to 224 × 224 pixels. An example batch of the PlantVillage dataset is shown in Fig.~\ref{fig: plant_village_batch}. We perform both the model optimization and predictions on these downscaled images.

\textit{Approach:}
We perform experiments with three ImageNet-pretrained models - VGG-16, ResNet-50, and EfficientNetB0. These models have varying numbers of parameters, sizes, and performance on the ImageNet dataset. It has been shown that these pre-trained models perform better than a model trained from scratch on the PlantVillage dataset.

During training, we use categorical cross-entropy loss with Adam optimization method with an initial learning rate of 2e-5 , the beta values of (0.9, 0.999) and an epsilon as 1e-08. A low learning rate is used to prevent the divergence of the model and to preserve primitive image filters identified during pre-training. The batch size used during training is 32, and the number of epochs used is 25. In addition, we also use early stopping and model checkpointing based on the validation loss, which gives us the best performing model on the validation dataset.

\begin{figure}[htbp]
\centerline{\includegraphics[width=0.5\textwidth]{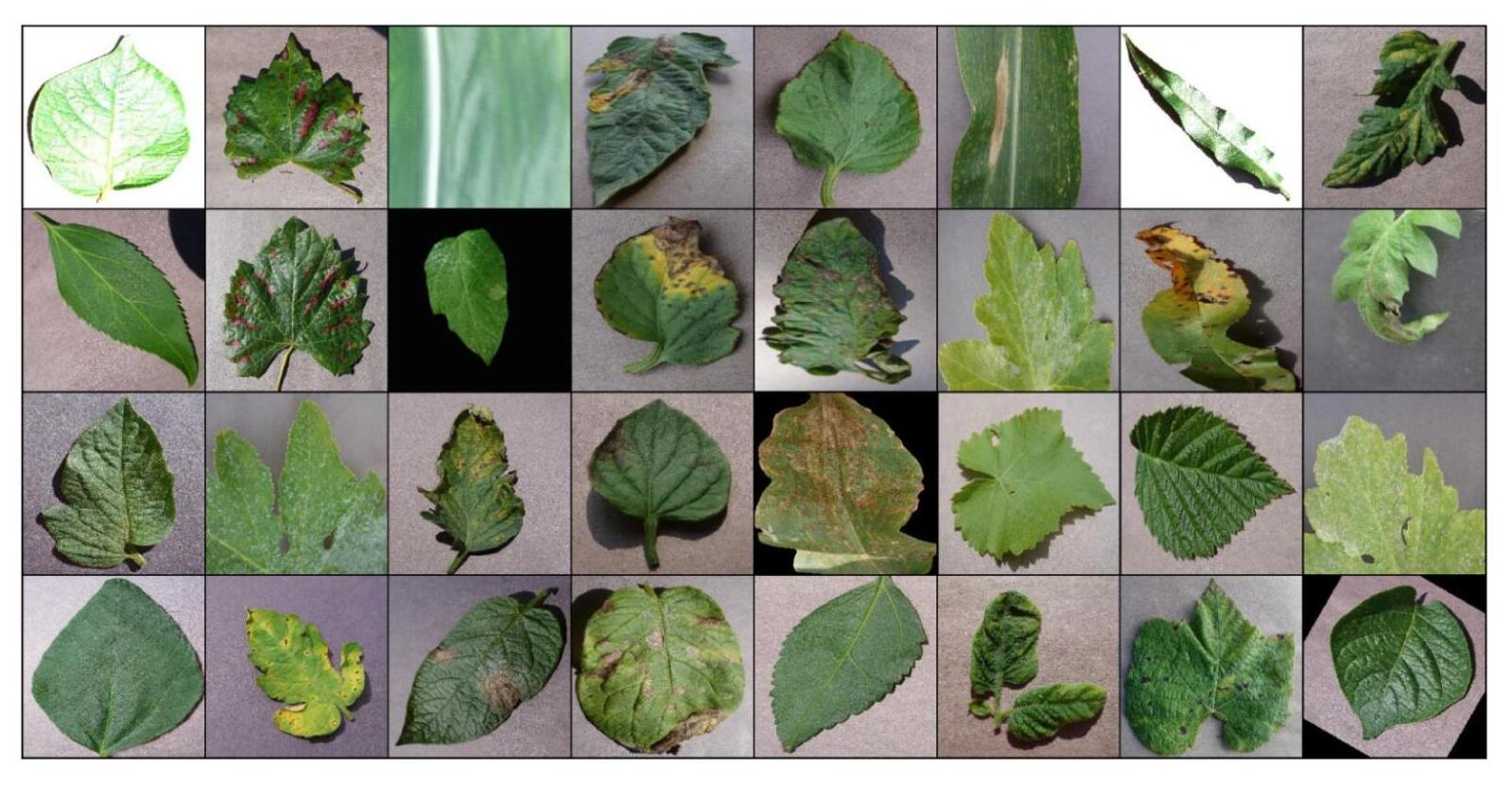}}
\caption{An example batch of the PlantVillage Dataset
}
\label{fig: plant_village_batch}
\end{figure}

We also note down the accuracy of the model during the training. The performance of these models may be further improved with training.

It is advisable to use GPU instead of CPU when dealing with images dataset because GPUs allow several parallel computations, thereby facilitating faster training and inference. We use the free GPUs provided by Kaggle and Google Colab for our experiments.

Lastly, we use the LIME method in order to understand the predictions made by our best model. For this, we use 1000 samples in the LIME method from the lime package, and we check the positively weighted segments towards the predicted class. Then, we plot the top-10 important segments on the image and display it on the application. Due to computation restrictions, however, we only use 249 samples on the application. Larger numbers of samples lead to better explanations with LIME.
\subsubsection{Fertilizer Recommendation}
\textit{Dataset Description:} For Fertilizer recommendation, the dataset we use is a custom dataset\footnote{Fertilizer Recommendation Dataset: \url{https://github.com/Gladiator07/Harvestify/tree/master/Data-raw}} with 5 features– Crop, N, P, K, pH, and soil moisture. There are 22 crops such as rice, maize, coffee beans, etc. with their respective ideal N,P, and K values. The data set represents the value of N,P and K the soil should have in order for the crop to grow in the most efficient way. Depending on the deficiency of the N, P or K value a fertilizer is recommended to the farmer.
\\
\textit{Approach:} We have used rule-based classification– a classification scheme that makes use of IF-THEN rules for class prediction –  to provide the best fertilizer for a plant. Depending on how far a plant is from its ideal N, P, or K value a fertilizer is recommended. For our purposes, we have 6 types of fertilizer recommendations currently, based on whether the N/P/K values are high or low.
\section{Results and Discussion}

\subsection{Crop Recommendation}
The results for our crop recommendation experiments are shown in the Table~\ref{tab: accuracy_crop_recommendation}. The Fig.~\ref{fig: acc_compare_crop_reco} also depicts these scores on a bar-chart for easy comparison. We can see that the RandomForest and NaiveBayes models perform the best, followed by the XGBoost model. It is expected that boosting (RandomForest) and bagging (XGBoost) models will usually perform and generalize better than non-ensemble methods.

\begin{table}[htbp]
\caption{Accuracy Comparison of Crop Recommendation models}
\begin{center}
\begin{tabular}{|c|c|}
\hline
\textbf{Model Type} & \textbf{\textit{5-Fold Cross-Val Accuracy}} \\
\hline
Decision Tree& 0.914\\
\hline
Naive Bayes& 0.995\\
\hline
SVM& 0.983\\
\hline
Logistic Regression& 0.955\\
\hline
Random Forest& 0.995\\
\hline
XGBoost& 0.992\\
\hline
\end{tabular}
\label{tab: accuracy_crop_recommendation}
\end{center}
\end{table}

\begin{figure}[htbp]
\centerline{\includegraphics[width=0.5\textwidth]{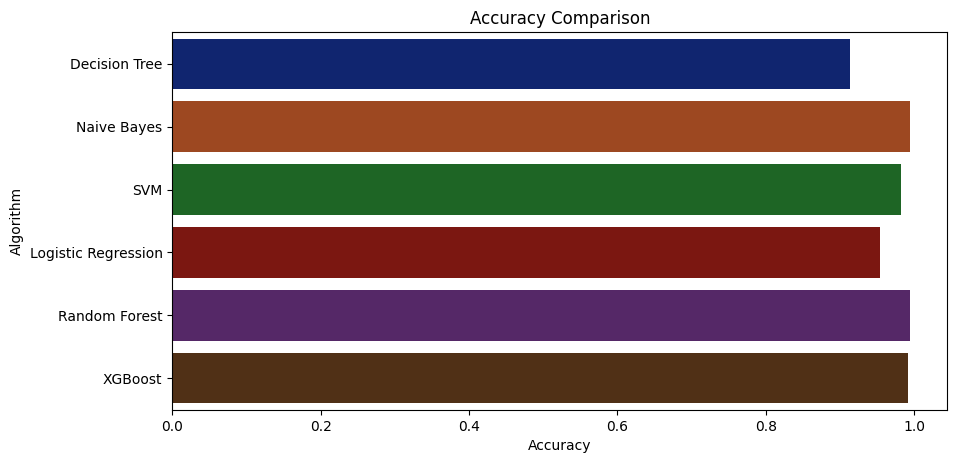}}
\caption{Accuracy Comparison of Crop Recommendation models
}
\label{fig: acc_compare_crop_reco}
\end{figure}

\begin{figure}[htbp]
\centerline{\includegraphics[width=0.5\textwidth]{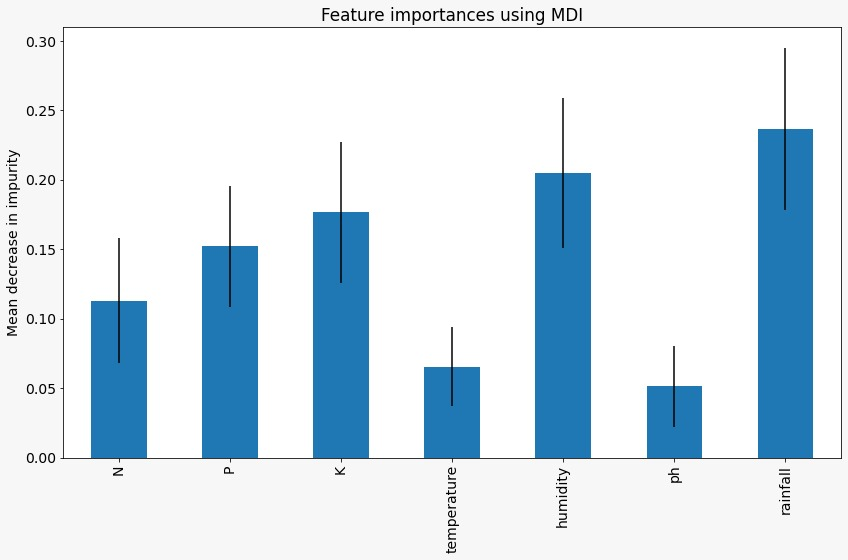}}
\caption{Feature importance for Crop Recommendation RandomForest model}
\label{fig: feature_importance_crop_reco}
\end{figure}

We choose the RandomForest model, which has a cross-validation accuracy of 0.995 for our application because we are able to easily understand the feature importances of the feature used, which tells us how important the features are for our classification.

The Fig.~\ref{fig: feature_importance_crop_reco} depicts the feature importance we calculate from the RandomForest model. We observe that rainfall plays the most significant role in determining the crop type. The second highest importance is that of humidity, followed by K, P and N. This means that overall water content is of the highest importance, followed by soil quality. Thus, using this model, we are able to also understand which features are overall important to our model for crop recommendation.

\subsection{Crop Disease Detection}
After getting the optimal validation scores for VGG, ResNet and EfficientNet, we conclude that the EfficientNet model performs the best of all. Fig.~\ref{fig: acc_compare_plant_disease} shows the accuracy comparison for the three models, where this is apparent. We also plot accuracy and loss curves for the three models in Fig.~\ref{val_loss_curve_pdd}. From the accuracy curves, we observe that EfficientNet reaches the highest score very soon, compared to the other two models. ResNet reaches a score close to EfficientNet, but 
\begin{figure}[htbp]
\centerline{\includegraphics[width=0.5\textwidth]{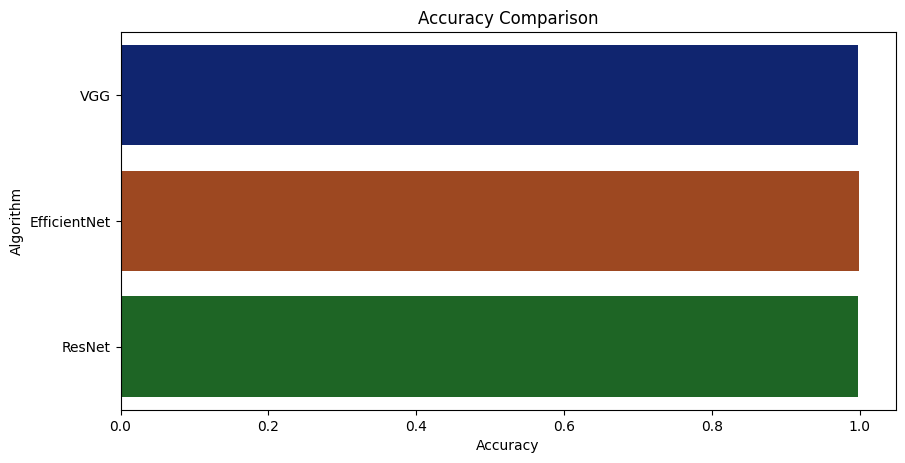}}
\caption{Accuracy Comparison of Plant Disease Detection models
}
\label{fig: acc_compare_plant_disease}
\end{figure}

\begin{figure}[htbp]
\centerline{\includegraphics[width=0.5\textwidth]{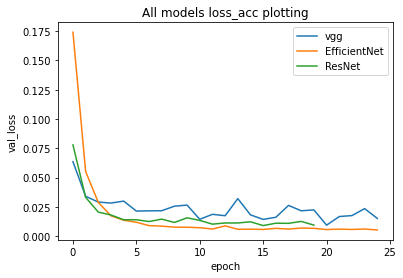}}
\caption{Validation Loss Curves for Plant Disease Detection models}
\label{fig: val_loss_curve_pdd}
\end{figure}
much later in its training. VGG is a smaller model, which could be why it fails to learn the data well and does not perform as well as the other two models. The EfficientNet is a class of several state-of-the-art convolutional neural network based models which are created using a structured approach for scaling neural networks. They have been pretrained on the ImageNet dataset, and achieve the highest score of all the previously designed CNN-based models on ImageNet.

For these reasons, we choose the EfficientNet model and deploy it on our application to perform leaf image classification.

\subsection{Interpretability Analysis}

\subsubsection{LIME}
This EfficientNet model is then used when computing the LIME explanation for the samples. For each sample, LIME produces an image with the respective important segments highlighted in the image. This tells us what the model is focusing on in the image to make its prediction.

In the Fig.~\ref{fig: lime_rust_3} and Fig.~\ref{fig: lime_rust_2}, we see that the diseased regions are marked as positive segments, which suggests the regions which the model looks at are consistent with the diseases portions of the leaf. With a higher number of samples, we expect to see better explanations, which are fine-grained. In addition, we plan to increase the image size and use better segmentation algorithms to be able to generate better and more fine-grained explanations in a future work. This helps in understanding whether the model looks at the correctly diseased regions or not, and which kinds of images may be further added to the training dataset to improve the generalization score.

\begin{figure}
     \centering
     \begin{subfigure}[b]{0.5\textwidth}
         \centering
         \includegraphics[width=0.5\textwidth]{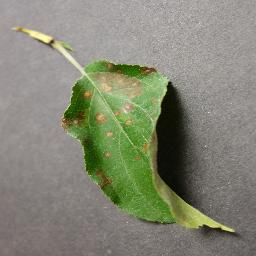}
         \caption{Leaf Image Input}
         \label{fig: lime_rust_3_in}
     \end{subfigure}
     \hfill
     \begin{subfigure}[b]{0.5\textwidth}
         \centering
         \includegraphics[width=0.5\textwidth]{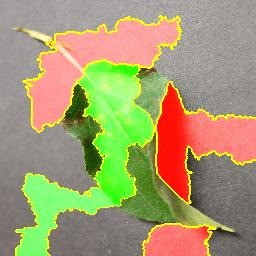}
         \caption{LIME Output}
         \label{fig: lime_rust_3_out}
     \end{subfigure}
        \caption{An example of LIME explanation method on AppleCedarRust3 leaf}
        \label{fig: lime_rust_3}
\end{figure}

\begin{figure}
     \centering
     \begin{subfigure}[b]{0.5\textwidth}
         \centering
         \includegraphics[width=0.5\textwidth]{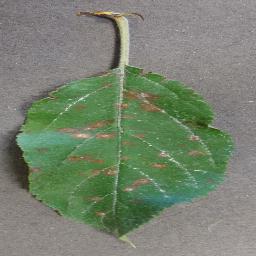}
         \caption{Leaf Image Input}
         \label{fig: lime_rust_2_in}
     \end{subfigure}
     \hfill
     \begin{subfigure}[b]{0.5\textwidth}
         \centering
         \includegraphics[width=0.5\textwidth]{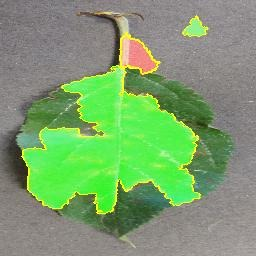}
         \caption{LIME Output}
         \label{fig: lime_rust_2_out}
     \end{subfigure}
        \caption{An example of LIME explanation method on AppleCedarRust2 leaf}
        \label{fig: lime_rust_2}
\end{figure}


\section{Conclusion and Future Work}
In this paper, we propose a user-friendly web application system based on machine learning and web-scraping called the ‘Farmer’s Assistant’. With our system, we are successfully able to provide several features - crop recommendation using Random Forest algorithm, fertilizer recommendation using a rule based classification system, and crop disease detection using EfficientNet model on leaf images. The user can provide the input using forms on our user interface and quickly get their results. In addition, we also use the LIME interpretability method to explain our predictions on the disease detection image, which can potentially help understand why our model predicts what it predicts, and improve the datasets and models using this information.

While our application runs very smoothly, we have several directions in which we can improve our application. Firstly, for crop recommendation and fertilizer recommendation, we can provide the availability of the same on the popular shopping websites, and possibly allow users to buy the crops and fertilizers directly from our application.

Another improvement that can be done with fertilizer recommendation is that we want to be able to find data on various brands and items available based on the N,P,K values. Currently, we only provide six kinds of recommendations, but in future, we want to be able to use complex machine learning systems to provide finer recommendations.

Next, we understand that the dataset we have used for disease classification is not exhaustive. This means that our model performs well only on the images which are from the classes the model already knows. It will not be able to detect the correct class for any out-of-domain data. This problem needs to be addressed in the future, and there are two ways of going about this. One solution is to find other datasets of similar scales with other types of crops and/or diseases, or generate/scale those datasets using generative modeling, and then add them to our training set. This will allow our model to generalize better. The second option is to allow the users to input their own images by creating a portal on our web-application to annotate the images themselves.

It has also been shown that LIME explanations in themselves are not always reliable as they only provide local information about an example, and not what the model focuses on at a global level. Hence, we can use other methods like GradCAM, Integrated Gradients, etc.  or other training approaches like sparse-linear layers with LIME in order to explain our model predictions better.

Lastly, we also wish to provide fine-grained segmentations of the diseased portion of the dataset. Currently, this is not possible due to lack of such data. However, in our application, we can integrate a segmentation annotation tool where the users might be able to help us with the lack. Also, we can use some unsupervised algorithms to pin-point the diseased areas in the image. We intend to add these features and fix these gaps in our upcoming work.

\end{document}